\pdfoutput=1
\documentclass[11pt]{article}

\usepackage[final]{acl}

\usepackage{times}
\usepackage{latexsym}

\usepackage[T1]{fontenc}
\usepackage[utf8]{inputenc}

\usepackage{microtype}

\usepackage{inconsolata}

\usepackage{graphicx}

\usepackage{algorithm}
\usepackage{algorithmic}

\usepackage{booktabs, multirow} % For formal tables
\usepackage{xspace}
\usepackage{color, xcolor,colortbl} 
\usepackage{amsmath}
\usepackage{tikz}
\usepackage{subfigure}

\newcommand{\model}{ReasVQA\xspace}

\newcommand{\xj}[1]{\textcolor{black}{#1}} % change red red to black to merge all
\newcommand{\ljx}[1]{\textcolor{black}{#1}}

\usepackage{pifont}

\title{ReasVQA: Advancing VideoQA with Imperfect Reasoning Process}

\author{Jianxin Liang$^{1}$ , Xiaojun Meng$^{2}$ , Huishuai Zhang$^{1}$\thanks{Corresponding author.} , \\
\textbf{Yueqian Wang$^{1}$ , Jiansheng Wei$^{2}$ , Dongyan Zhao$^{1,3}$\thanks{Corresponding author.} } \\
  $^1$ Wangxuan Institute of Computer Technology, Peking University \\
  $^2$  Huawei Noah’s Ark Lab \\
  $^3$ National Key Laboratory of General Artificial Intelligence, Peking University  \\
  \texttt{\{liangjx, wangyueqian, zhanghuishuai, zhaody\}@pku.edu.cn}, \\
  \{xiaojun.meng, weijiansheng\}@huawei.com \\
  }

\begin{document}
\maketitle

\begin{abstract}

Video Question Answering (VideoQA) is a challenging task that requires understanding complex visual and temporal relationships within videos to answer questions accurately. In this work, we introduce \textbf{ReasVQA} (Reasoning-enhanced Video Question Answering), a novel approach that leverages reasoning processes generated by Multimodal Large Language Models (MLLMs) to improve the performance of VideoQA models. Our approach consists of \xj{three phases: reasoning generation, reasoning refinement, and learning from reasoning. First, we generate detailed reasoning processes using additional MLLMs, and second refine them via a filtering step to ensure data quality. Finally, we use the reasoning data, which might be in an imperfect form, to guide the VideoQA model via multi-task learning, on how to interpret and answer questions based on a given video.}
\xj{We evaluate ReasVQA on three popular benchmarks, and our results establish new state-of-the-art performance with significant improvements of +2.9 on NExT-QA, +7.3 on STAR, and +5.9 on IntentQA. Our findings demonstrate the supervising benefits of integrating reasoning processes into VideoQA. Further studies validate each component of our method, also with different backbones and MLLMs, and again highlight the advantages of this simple but effective method. We offer a new perspective on enhancing VideoQA performance by utilizing advanced reasoning techniques, setting a new benchmark in this research field.}

% \huzhangc{\textbf{General comments about the paper, not specific to abstract.} 

% 1. why do we care about the performance of \textbf{smaller} VideoQA models?  If we want SOTA performance on these tasks, it is obvious to use extreme large models rather than 3B or 7B models. Perhaps say that small model is essential to deploy on the edge devices. and videoQA is one of the most important tasks for real-world robots.

% 2. why do we use \textbf{InternVL} as reasoning generator? I saw some argument that ``is one of SOTA and open-source MLLMs, 26B version can be deploy on a single H800 GPU'', which is not very persuasive. Why not commercial models? Perhaps, the licence issues? please check if commercial models are allowed to do data distillation. Why not even larger models? Perhaps, not open-sourced and not available in service. We should have a more comprehensive overview about the MLLMs and justify our choice.

% 3. We may emphasize that InternVL demonstrates weaker performance than small models, which, at first glance, indicates that there is no possibility to conduct model distillation from InternVL to small specific models. In general the approach in the paper fall into the ``model distillation via data'' category and Liu Chang has presented several recent works along this line?  
% }

\end{abstract}

% TODO: 
% 1. 语法问题；
% 2. 时态前后统一：一般现在时；
% 3. 全文用词的统一：
% 4. 其他存在的问题

\section{Introduction}

Video Question Answering (VideoQA)~\cite{patel2021recent, zhong2022Video} is an increasingly important task within the fields of artificial intelligence and computer vision, aiming to enable machines to understand and answer questions about video content. 
% This capability is crucial for a wide range of applications, including autonomous systems, educational technologies, content recommendation, and interactive media. 
% \input{2_tab/internvl26}
It poses unique challenges due to the complex nature of video data~\cite{zhong2022Video}, which combines temporal and spatial information, often requiring deep contextual understanding and reasoning over sequences of frames.

Existing approaches to VideoQA~\cite{pan2023retrieving, liang2024end, wang2024lstp, yu2023self} typically involve models that attempt to map video frames and questions directly to answers. While these methods have shown some success, they often fall short when complex reasoning or temporal relationships are involved~\cite{mangalam2023egoschema, kahatapitiya2024language}. These methods lack the ability to explicitly break down complex visual content into manageable components, which hinders the model’s capacity for deeper comprehension and accurate responses. 
Meanwhile, process supervision~\cite{lightman2023let, uesato2022solving} has been shown to be comparable to, or even outperform, outcome supervision in certain mathematical scenarios. Building on this insight, we try to enhance the model's performance in VideoQA by focusing on improving its advanced reasoning capabilities in this work.
However, obtaining accurate and well-annotated reasoning processes, especially through human annotation, is both costly and time-consuming, making it challenging to scale this approach effectively.

On the other hand, Multimodal Large Language Models (MLLMs)~\cite{team2023gemini, achiam2023gpt,chen2024far} have demonstrated impressive capabilities in generating detailed explanations, captions, and even reasoning processes for both images and videos, achieving SOTA across many tasks. These models also exhibit strong multimodal understanding~\cite{lin2023video,li2024llava,zhang2023multimodal}, capable of answering questions about unseen content, thus opening up new avenues for improving VideoQA. Leveraging their powerful multimodal chat abilities, MLLMs have shown the potential to automatically generate synthetic reasoning data, offering a way to bypass the need for costly human annotations.

However, directly leveraging these synthetic reasoning processes presents significant challenges. While MLLMs can generate detailed and often insightful reasoning, their outputs are not always dependable. Due to inherent biases or occasional misunderstandings, these models may produce incorrect or irrelevant reasoning. For example, as shown in Table~\ref{tb:internvl}, a SOTA MLLM, InternVL~\cite{chen2024internvl,chen2024far}, demonstrates an accuracy of less than 70\% across three VideoQA datasets when generating reasoning processes and final answers—leaving over 30\% of outputs erroneous. Naturally, such reasoning processes may contain critical mistakes. Directly feeding this flawed synthetic reasoning into VideoQA models can hinder performance, as the inaccuracies may propagate throughout the learning process. This raises a crucial question: \ljx{How to address the inconsistency between these imperfect reasoning steps and the true answers, and  leverage them effectively during training?}
% Can we eliminate these errors, and if so, how can we effectively use them?
% Is there a better way to harness these imperfect synthetic reasoning processes to enhance VideoQA performance?

\begin{figure*}[!thb]
    \centering
    \includegraphics[scale=0.45]{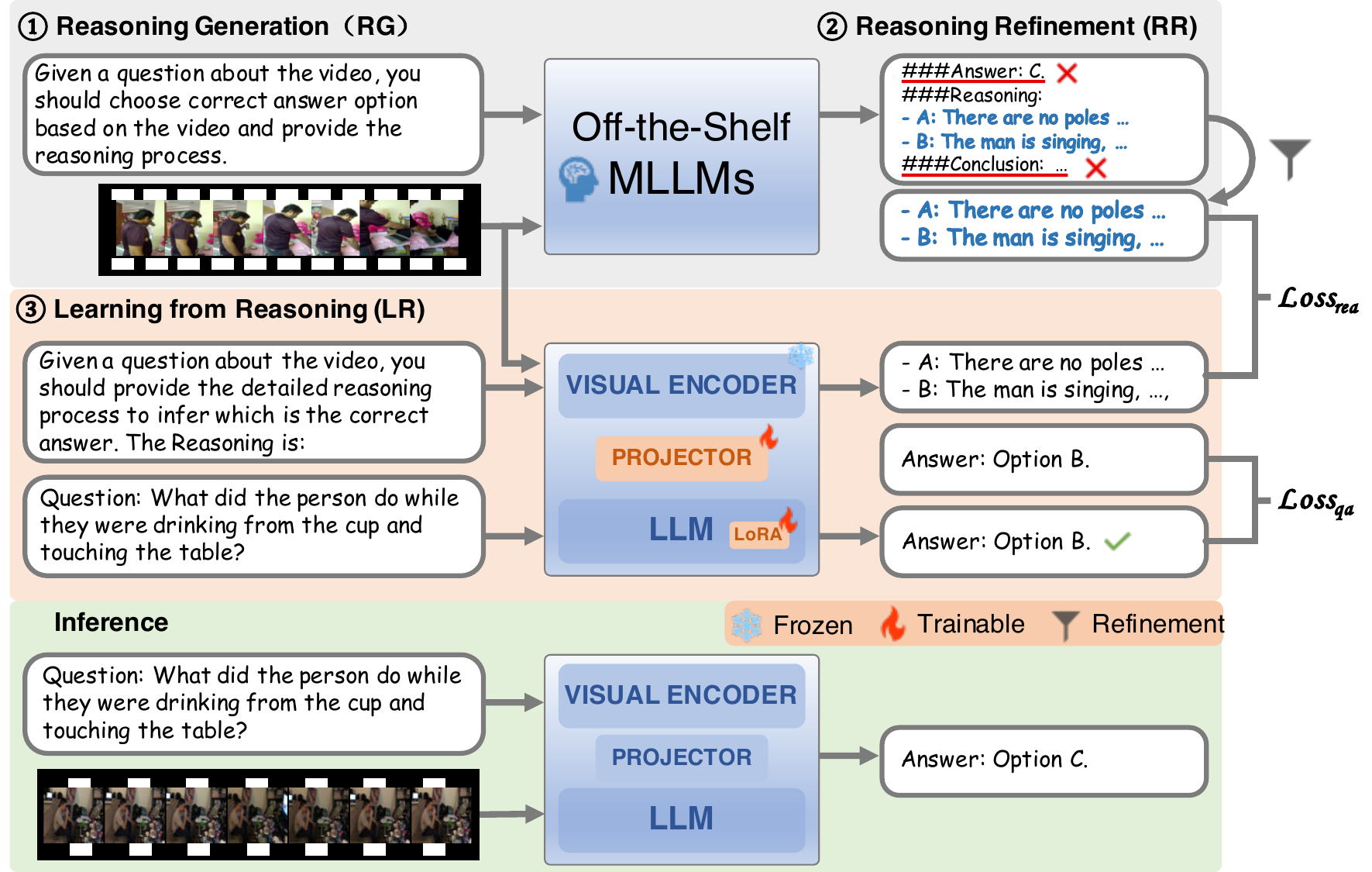}

    \caption{Overview of our method \model. \textcircled{\scriptsize{1}}~Reasoning Generation: a SOTA MLLM is prompted to solve complex questions by generating detailed reasoning explanations. \textcircled{\scriptsize{2}}~Reasoning Refinement: we process and refine the reasoning steps to alleviate the conflict with true answers. \textcircled{\scriptsize{3}}~\xj{Learning from  Reasoning (Multi-task Training): the refined reasoning steps are used to guide a model to improve its performance on the VideoQA tasks.}}
    \label{fig:overview_arch}
    % \vspace{-4mm}

\end{figure*}

%To alleviate the above problems,
% In this paper, 
We introduce \textbf{\model} (Reasoning-enhanced VideoQA), a novel approach that leverages the reasoning processes generated by MLLMs to improve the performance of VideoQA models. Our approach comprises three phases: Reasoning Generation (\textbf{RG}), Reasoning Refinement (\textbf{RR}), and Learning from Reasoning (\textbf{LR}), as shown in Figure~\ref{fig:overview_arch}. 
In the \textbf{RG} phase, we utilize SOTA MLLMs to produce reasoning processes for a set of VideoQA tasks, which are then utilized in the \textbf{LR} phase to train the actual model. Based on the outputs of the MLLMs, we evaluate the correctness of these reasoning processes by examining the associated answers, as illustrated in Figure~\ref{fig:example}. To mitigate the impact of potential errors within these reasoning processes, we apply data filtering to clean and refine the generated outputs during \textbf{RR} phase. Specifically, we remove the sentences containing  answers, shifting the focus towards the reasoning steps rather than the conclusions. When the generated reasoning steps contains imperfections, these filtered reasoning steps still offer valuable insights for the model. 
The third phase, \textbf{LR}, employs a multi-task learning framework, where the VideoQA model is trained to both answer questions and generate reasoning processes simultaneously. The training supervision utilizes the dataset's original true answer annotations alongside the refined reasoning processes generated during the RG phase, ensuring that the model learns from both correct answers and the cleaned, structured reasoning paths. This phase allows the model not only to improve its question-answering accuracy but also to develop the ability to articulate the reasoning behind its answers. By integrating these curated reasoning processes into the training regimen, \model can inherit and refine the reasoning capabilities seen in larger MLLMs, leading to a more robust understanding of video.

We validate the effectiveness of our approach through extensive experiments across multiple model architectures and datasets. Experiments demonstrate significant improvements, achieving new state-of-the-art results with increases of +2.9 on NExT-QA, +7.3 on STAR, and +5.9 on IntentQA~\cite{xiao2021next,wu2021star,li2023intentqa}. These findings validate the integration of generated reasoning processes into VideoQA models. Moreover, our detailed analyses confirm the effectiveness of each phase of the approach, offering insights into how reasoning refinement and multi-task learning contribute to the overall performance improvements.

In summary, the contributions of this paper are:

\xj{1. we propose ReasVQA to demonstrate the potential of using even imperfect reasoning processes from additional MLLMs to guide VideoQA models and also show how refined reasoning data can lead to significant performance gains.}

\xj{2. we introduce a multi-task training method that incorporates reasoning into VideoQA tasks, offering a new perspective on learning from reasoning and knowledge integration.}

\xj{3. we provide empirical evidence of our approach's effectiveness via rigorous experiments and analyses, setting new SOTAs in the VideoQA field.}

% 模型整体流程架构

\section{Related Work}

\paragraph{Video Question Answering} 
% To drive the community forward and comprehensively assess model capabilities, researchers have constructed VideoQA-related datasets~\cite{tapaswi2016movieqa, xu2017video,lei2018tvqa,zhao2018open,xiao2021next,wu2021star}. 
% These datasets cover multiple aspects including video content, spatiotemporal understanding, and causal reasoning, significantly advancing the field of VideoQA. 
VideoQA typically requires models to comprehend dynamic scenes, temporal information, and multimodal cues (such as visual, audio, and text) from videos. To better capture these features and understand their interactions across modalities, VideoQA methods have evolved significantly from earlier approaches like attention mechanisms, memory modules, and graph neural networks to leveraging more advanced pre-trained models~\cite{xu2017video, jang2017tgif,khan2020mmft,yang2020bert,yang2021just,lei2021less,wang2022internvideo,ye2023hitea,gao2023mist,wang2023all,ijcai2023p582}. For example, InternVideo~\cite{wang2022internvideo} extends a vision transformer pre-trained on images for video representation learning. Recently, with the growing capabilities of foundation models, some studies~\cite{yu2023self,liang2024end, wang2024lstp, fei2024video} have aimed to tackle VideoQA tasks by using large language models (LLMs)\cite{ko2023large,liu2024visual, li2024llava}. For instance,~\citet{wang2023vamos} and~\citet{ko2023large} try to leverage LLaMA’s~\citep{touvron2023llama} knowledge of temporal and causal reasoning to address the complexities of VideoQA. However, these approaches typically rely solely on supervising the model using the final result, which often falls short or leads to overfitting, especially when complex reasoning or temporal relationships are involved.

\paragraph{Learning by using LLMs/MLLMs}
Recently, LLMs have demonstrated strong reasoning abilities~\cite{wei2022chain, kojima2022large, achiam2023gpt, chowdhery2023palm, brown2020language, wang2022self}, effectively solving multi-step reasoning tasks by explicitly performing intermediate reasoning steps using Chain-of-Thought or few-shot prompts. Furthermore, \citet{uesato2022solving} and \citet{lightman2023let} find that process supervision, where intermediate reasoning steps are supervised, can yield performance that is comparable to or even better than final result supervision in certain mathematical scenarios. This provides a promising direction for integrating reasoning processes into VideoQA models in our work.

% Building on this strength, some works have used these reasoning steps to enhance input prompts~\cite{wang2022self} or as additional fine-tuning data~\cite{hsieh2023distilling}, thereby improving LLMs' performance.

Inspired by the successes of LLMs, MLLMs have also explored similar strategies~\cite{guo2023images,zeng2022socratic,li2023videochat,zhang2023simple,romero2024question,fei2024video}. 
For instance, models like 
% Img2LLM~\cite{guo2023images},  incorporate additional linguistic information, such as image captions, to answer visual questions. In video-language tasks, methods like Socratic Models~\cite{zeng2022socratic} and VideoChat~\cite{li2023videochat} integrate pretrained visual models with LLMs for extracting visual concepts and applying them to video tasks. 
LLoVi~\cite{zhang2023simple}, utilizes GPTs to generate visual descriptions or summaries from captions. Q-ViD~\cite{romero2024question} enhances zero-shot video understanding by incorporating captions relevant to the questions into the model's input. MotionEpic\cite{fei2024video} breaks down the raw intricate video reasoning problem into a chain of simpler sub-problems and solves them one by one. 
% sequentially. 

% We avoid using few-shot approaches in this context because they require a strong foundational capability in VideoQA, including the ability to follow instructions and produce outputs in a specified format. In our scenarios, the quality of the provided reasoning becomes crucial, especially given our current setting. Imperfect reasoning further complicates the task for the base model, making it even more challenging to perform well.

Unlike the aforementioned MLLM-based works that primarily rely on final result supervision, in this paper, we focus on mitigating the negative impact of errors in reasoning processes and explore different forms of supervision that can be used for training models, particularly process supervision, which leverages reasoning as a supervision signal. While this approach has been discussed and utilized in the context of LLMs, to the best of our knowledge, this is the first work to introduce process supervision into VideoQA. To achieve this, we decouple the reasoning steps from their predicted answers or conclusions through a simple refinement process, which allows the model to focus more on the reasoning steps during training, rather than the given predictions no matter correct or not. In the subsequent training phase, we employ a multi-task learning approach, where the model learns to both perform the VideoQA task and regenerate the reasoning process. We believe this enables the model to perform videoQA tasks following a logical manner and thus get better performance.
%extract useful information for QA tasks, ultimately enhancing its performance on the VideoQA task.
%\xjc{should be related to reasoning but not just simple information extraction.}

\section{Methodology}
To harness imperfect synthetic reasoning data to guide VideoQA, 
we propose the three-phase \textbf{ReasVQA} involving Reasoning Generation, Reasoning Refinement, and Learning from Reasoning.

\subsection{Reasoning Generation}
\label{sec:rg}

We utilize existing multimodal large language models (MLLMs) as reasoning process generators to produce a reasoning process $r$ for a given video $v$ and question $q$, as illustrated in Figure~\ref{fig:example}. 

%Therefore, the correctness of the reasoning steps cannot be guaranteed. Some reasoning steps may contain unavoidable errors that compromise the accuracy of outputs. 
% \xj{However, due to inherent limitations of MLLMs particularly in a zero-shot setting},
\ljx{However, even SOTA models still regularly produce logical errors, especially in zero-shot settings, i.e., }
the predicted answer $\hat{a}$ included in $r$ is sometimes incorrect. To assess the quality of the generated reasoning processes, we evaluate them based solely on the correctness of their predicted answers.
% ~\xjc{very important step, need to hightlight and mention how to do it}
Specifically, we compare \(\hat{a}\) with the ground-truth labels \(a\). If \(\hat{a}\) does not equal \(a\), we consider the reasoning process \(r\) to be imperfect, denoting it as \textbf{Incorrect Reasoning}; conversely, if they match, \(r\) is classified as \textbf{Correct Reasoning}.

For example, as illustrated in Figure~\ref{fig:example}, for the question "\textit{Which object was tidied up by the person?}", the answer $\hat{a}$ is "\textit{The closet/cabinet}" while "\textit{The blanket}" is the correct answer. \ljx{Therefore, this reasoning is deemed imperfect and classified as Incorrect Reasoning.}
\ljx{This evaluation helps to assess the quality of reasoning processes generated by MLLMs and provides guidance for the next steps to better refine these imperfect reasoning processes}.

\subsection{Reasoning Refinement}
%Given the imperfect reasoning process, especially the final answers sometimes are wrong, xxx

As shown in Table~\ref{tb:internvl}, the accuracy of MLLMs is significantly lower than the current SOTAs in VideoQA tasks. This suggests that over 30\% of the generated reasoning is unreliable. On the other hand, with a close examination of the incorrect answers generated by MLLMs, we find that the reasoning steps are often valuable although the final answers are wrong. Therefore, we try to decouple the reasoning from the final predicted answer. This leads to the following refinement process: we retain only the essential steps that do not include any conclusion through keyword matching, regardless of whether they are classified as \textit{Correct Reasoning} or \textit{Incorrect Reasoning}. For \textit{Incorrect Reasoning}, we further remove any words that contain the ground truth, which allows the model to focus on the logical flow of the reasoning.

% \xj{To mitigate the potential impact of errors in the reasoning steps $r$ for training, we further process and clean the reasoning data. We only retain the essential reasoning steps $\hat{r}$ that do not include any answer, regardless it is correct or not. 
% The key objective is to decouple the reasoning from the final predicted answer, which is given by an MLLM and indeed often wrong as shown in Table~\ref{tb:internvl}.
% We achieve this by simply removing sentences in the reasoning that contain explicit answers or conclusions through keyword matching, allowing the model to focus on the logical flow of the reasoning.
% ~\xjc{need to describe how to perform it, and how is the accuracy of this cleaning process? Can we form it as a standard methodology? If so, we can create a new section named as 3.2 Data cleaning for it.} 

%By discarding the MLLMs' final prediction or conclusions, we retain the essential reasoning steps $\hat{r}$ that contain the remaining options, allowing the model to learn from the logical flow even if the predicted answer is incorrect, which is also the reason that we call it imperfect reasoning process.
This refinement is simple yet effective. 
We take a refined example in Figure~\ref{fig:example}, "\textit{The individual is standing in front of a wooden cabinet with slatted doors}" provides relevant context for the video, while "\textit{The other objects mentioned in the hints, such as the table and clothes, are not visible or being interacted with}" \xj{uses the logical process of elimination to exclude incorrect options. Both reasoning steps remain valuable, even if the final predicted answer is wrong and the sentences might be incomplete after refinement. This is also the reason we call it an imperfect reasoning process.}
%by refining these reasoning processes, we extract the underlying logical aids from a teacher, despite potential inaccuracies.

\xj{Overall, this refinement enables the VideoQA model to particularly focus on the structure of reasoning, filtering out erroneous information or shortcuts. It allows the model to strengthen the ability to generate coherent reasoning in response to video-based questions, improving performance across various VideoQA tasks. Importantly, this entire process occurs solely in the training phase, ensuring that no information leakage occurs in evaluation. }
% ~\xjc{To my undestanding, this learing from reasoning is only performed on train set instead of test set right? We need to highlight it somewhere in the paper. It is better if we can prove that there is no information leakage.}

\subsection{Learning from Reasoning}\label{sec:training}
The third phase of our approach aims to effectively transfer the valuable information from refined reasoning processes $\hat{r}$ to a VideoQA model $f(\cdot)$. To accomplish this, we explore different training approaches, such as single-task learning (STL) and multi-task learning (MTL).

In the STL approach, we concatenate the refined reasoning $\hat{r}$ with the ground-truth answer $a$, denoted as $\hat{r}_a$, and use this joint text as the supervision signal for the model. The model is trained to generate both the reasoning and the correct answer sequentially. The objective function is:
\begin{equation}
\mathcal{L}_{st} = \mathcal{C}(f(v,q), \hat{r}_a).
\end{equation}\label{eq:loss_st}
where $\mathcal{C}(\cdot)$ denotes the cross-entropy loss. However, this method proves challenging because, even after refinement, there may still be inconsistencies within $\hat{r}_a$, \xj{making it difficult for the model to learn as well as reconcile discrepancies.}

\ljx{An alternative, more effective method is the MTL approach. In this way, the VideoQA model $f(\cdot)$ learns to simultaneously perform the VideoQA task and reconstruct the reasoning process $\hat{r}$. To enhance flexibility in model training, we use a weighted sum of two loss functions}: 
\begin{equation}
\mathcal{L}_{mt} = \alpha *\mathcal{C}_{qa} (f(v,q), a) + \beta *\mathcal{C}_{rea}(f(v,q), \hat{r}) .
\end{equation}\label{eq:loss_mt}
where $\alpha$ and $\beta$ are weights such that $0 < \alpha, \beta < 1$ and $\alpha + \beta = 1$. \ljx{\(\mathcal{C}_{qa}(\cdot)\) and \(\mathcal{C}_{rea}(\cdot)\) represent the cross-entropy losses for QA and reasoning generation, respectively.}
By integrating reasoning generation as an auxiliary task, MTL allows the model to benefit not only from the direct supervision of the true answers but also from learning the logical reasoning processes, thus enhancing its understanding and reasoning capabilities in complex video scenarios. Moreover, adjusting the balance between \(\mathcal{C}_{qa}(\cdot)\) and \(\mathcal{C}_{rea}(\cdot)\) with different weights helps mitigate the propagation of any residual error in the reasoning, ensuring that the model can still learn effectively even with imperfect reasoning.

% TODO
% RG 过程具体用了哪些数据没说？
% 还有 Lora 也没具体说
% implement 提一下，系列可以放 appendix

\begin{table*}[!thb]
\centering
\resizebox{\linewidth}{!}{
\begin{tabular}{c|c|cccc|ccccc}
\toprule
\multirow{2}{*}{Model}  &\multirow{2}{*}{LLM Arch.} &\multicolumn{4}{c|}{NExT-QA}  &\multicolumn{5}{c}{STAR}  
% &\multirow{2}{*}{TVQA}   
\\
&  & Tem. & Cau. & Des. & Tot.\ $\uparrow$ & Int. & Seq. & Pre. & Fea. & Tot.\ $\uparrow$   \\ 
 
 \midrule

 \multicolumn{10}{l}{\textit{Non-LLM Models}}\\

Just Ask~\cite{yang2021just} &-   &51.4  &49.6  &63.1  &52.3  &-   &-   &-   &-     &-\\

All-in-One~\cite{wang2023all} &-     &48.6  &48.0  &63.2  &50.6  &47.5  &50.8  &47.7  &44.0  &48.9    \\

MIST~\cite{gao2023mist} & -  &56.6  &54.6  &66.9  &57.1  &55.5   &54.2   &54.2   &44.4   &54.0    \\

HiTeA~\cite{ye2023hitea} & -   &58.3  &62.4  &75.6  &63.1  &-  &-  &-  &-  &-     \\
InternVideo~\cite{wang2022internvideo} &  -   &58.5  &62.5  &75.8  &63.2  &62.7  &65.6  &54.9  &51.9  &58.7      \\
\midrule

\rowcolor{green!2} \multicolumn{11}{l}{\textit{LLM-based Models}} \\

\rowcolor{green!2} \color{gray} LLaMA-VQA~\cite{ko2023large} &\color{gray}LLaMA\ 7B &\color{gray}69.2  &\color{gray}72.7  &\color{gray}75.8 &\color{gray}72.0 
&\color{gray}66.2  &\color{gray}67.9  &\color{gray}57.2  &\color{gray}52.7  &\color{gray}65.4    \\
\rowcolor{green!2} \color{gray}MotionEpic~\cite{fei2024video} &\color{gray} Vicuna\ 7B   &\color{gray}74.6  &\color{gray}75.8  &\color{gray}83.3 &\color{gray}76.0 
&\color{gray}71.5  &\color{gray}72.6  &\color{gray}66.6  &\color{gray}62.7  &\color{gray}71.0   \\ 

\rowcolor{green!2} \color{gray}Vamos~\cite{wang2023vamos} &\color{gray} LLaMA2\ 7B   &\color{gray}72.3  &\color{gray}74.8  &\color{gray}81.6 &\color{gray}75.0 
&\color{gray}-  &\color{gray}-  &\color{gray}-  &\color{gray}-&\color{gray}-   \\

% InternVL &InternLM 20B   &63.9	&68.4	&68.4	&67  &67.3	&73.0	&67.9	&62.3	&69.8\\
% \rowcolor{green!2} BLIP-2$^{concat*}$ &FlanT5\ 3B   &68.1  &72.9  &81.2  &72.6  &65.4  &69.0  &59.7  &54.2  &62.0   \\

\rowcolor{green!2} LSTP~\cite{wang2024lstp} &FlanT5\ 3B  &66.5  &72.8  &81.2  &72.1  &-  &-  &-  &-  &-    \\

\rowcolor{green!2} SeViLA~\cite{yu2023self} &FlanT5\ 3B   &67.0  &73.8  &81.8 &73.8 
&66.4  &70.3  &61.2  &55.7  &67.2   \\

\rowcolor{green!02} VidF4~\cite{liang2024end} &FlanT5\ 3B   & 69.6 & 74.2 &83.3 & 74.1 & 68.4 & 70.4 & 60.9 & 59.4 & 68.1  
\\ 

\rowcolor{green!02}ViLA~\cite{wang2023vlap} & FlanT5\ 3B   &71.4  &73.6  &81.4 &74.1 
&70.0  &70.4  &61.2  &55.7  &67.2   \\

\midrule
\model(ours) & FlanT5\ 3B   & \textbf{73.0}  &\textbf{77.7}	&\textbf{82.8}
&\textbf{77.0} &\textbf{75.9}    &\textbf{76.6} &\textbf{67.3}   &\textbf{62.0} &	\textbf{74.5} 

\\ \bottomrule

\end{tabular}
}
\caption{\textbf{Model comparison on NExT-QA and STAR}. Specifically, Tem., Cau., Des., and Tot. denote Temporal, Causal, Description, and Total accuracy, respectively. Int., Seq., Pre., and Fea. denote Interaction, Sequence, Prediction, and Feasibility, respectively. 
}
\label{tb:main results}
\end{table*}

 % Just Ask \cite{yang2021just}, All-in-One \cite{wang2023all} and MIST \cite{gao2023mist}, HiTeA \cite{ye2023hitea} and InternVideo \cite{wang2022internvideo}. For LLM-based models, we use SOTA models such as BLIP-2 \cite{li2023blip}, LLaMA-VQA \cite{ko2023large}, LSTP \cite{wang2024lstp}, SeVILA \cite{yu2023self}, VidF4~\cite{liang2024end}, ViLA~\cite{wang2023vlap} and MotionEpic~\cite{fei2024video}.
\begin{table}[hbt]
\centering
\resizebox{\linewidth}{!}{

\begin{tabular}{c|c|cccc}
\toprule
\multirow{2}{*}{Model} &\multirow{2}{*}{LLM }  &\multicolumn{4}{c}{\textbf{IntentQA}}   \\

& & Why & How & Tem. & Tot. $\uparrow$   \\ 
 
 \midrule

HQGA~\cite{xiao2022video1} &-   &48.2  &54.3  &41.7  &47.7  \\
VGT~\cite{xiao2022video} & -  & 51.4  &56.0  &47.6  &51.3     \\
BlindGPT~\cite{ouyang2022training} &GPT3   & 52.2  &61.3  &43.4  &51.6      \\
CaVIR~\cite{li2023intentqa} &-   & 58.4  &65.5  &50.5  &57.6  \\
Vamos~\cite{wang2023vamos}  &LLaMA3 8B  & 69.5  &70.2  &65.0  &68.5     \\
MotionEpic~\citep{fei2024video} & Vicuna 7B  & -  &-  &-  &70.8     \\
LVNet~\cite{park2024too} &GPT-4o  &75.2  &71.6  &60.8  &71.1\\

\midrule
\model(ours) & FlanT5 3B  & \textbf{75.6}  &\textbf{93.3}	&\textbf{69.1}
&\textbf{77.0}  

\\ 
\bottomrule

\end{tabular}
}
\caption{{Model comparison on IntentQA.}
% \huzhangc{It is important to record the model size in the table.}
}
\label{tb:main results-intentqa}
\end{table}
% \begin{table}[!thb]
% \centering
% \resizebox{\linewidth}{!}{
% \begin{tabular}{c|lll|l}
% \toprule
% \multirow{2}{*}{Setting}  

% &\multirow{2}{*}{\textbf{NExT-QA}}  &\multirow{2}{*}{\textbf{STAR}}  &\multirow{2}{*}{\textbf{IntentQA}}  
% &\multirow{2}{*}{\textbf{Avg.\ \uparrow}}  \\
% & & &
% \\ 
 
%  \midrule
%  BLIP-FlanT5(3B) &74.5  &71.0  &73.0 & 72.8    \\

% w. our Rea  &77.0(+2.5) &74.5(+3.5) &77.0(+4.0) &76.2(+3.4)  \\
%  \midrule
% LLaVA-OV(0.5B) &64.2  &60.0  &58.9 &61.0   \\

% w. our Rea  &65.8(+1.6) &61.5(+1.5) &63.2(+4.3) &63.5(+2.5)   \\
% \midrule

% LLaVA-OV(7B) &77.5  &66.2  &74.1   &72.6 \\

% w. our Rea  &78.9(+1.4) &68.2(+2.0) &77.6(+3.5) &74.9(+2.3)  \\
% \midrule

% \end{tabular}
% }

% \caption{Adaptation of our method to different models.}
% \label{tb:llava5}
% \end{table}
\begin{table}[!thb]
\centering
\resizebox{\linewidth}{!}{
\begin{tabular}{c|ccc|l}
\toprule
\multirow{2}{*}{Setting}  

&\multirow{2}{*}{\textbf{NExT-QA}}  &\multirow{2}{*}{\textbf{STAR}}  &\multirow{2}{*}{\textbf{IntentQA}}  
&\multirow{2}{*}{\textbf{Avg.\ $\uparrow$}}  \\
& & &
\\ 
 
 \midrule
 BLIP-FlanT5(3B) &74.5  &71.0  &73.0 & 72.8    \\

w. \model  &77.0 &74.5 &77.0 &76.2(+3.4)  \\
 \midrule
LLaVA-OV(0.5B) &64.2  &60.0  &58.9 &61.0   \\

w. \model  &65.8 &61.5 &63.2 &63.5(+2.5)   \\
\midrule

LLaVA-OV(7B) &77.5  &66.2  &74.1   &72.6 \\

w. \model  &78.9 &68.2 &77.6 &74.9(+2.3)  \\
\bottomrule

\end{tabular}
}

\caption{ \model improves the performance of different models.}
\label{tb:llava5}
\end{table}

\section{Experiments}

In this section, we present our experiments on various VideoQA tasks. First, we describe the datasets we used and the implementation details. Then, we evaluate our method, compare \model with other SOTAs, and provide a comprehensive analysis.

\subsection{Datasets and Baselines}
\paragraph{Datasets and Evaluation Metrics}

We conduct experiments on three popular VideoQA datasets: NExT-QA, STAR, and IntentQA~\cite{xiao2021next, wu2021star, li2023intentqa}, which demand both causal and temporal reasoning abilities. 
% \paragraph{Evaluation Metrics.} 
For all these tasks, we employ the most used answer accuracy as the evaluation metric.  A higher accuracy score indicates better model performance. 

% See Appendix~\ref{app:data} for more details.

% \paragraph{Baselines.} We compare our method with two types of baselines: non-LLM and LLM-based models. For non-LLM methods, we use recent SOTA models, including Just Ask \cite{yang2021just}, All-in-One \cite{wang2023all} and MIST \cite{gao2023mist}, HiTeA \cite{ye2023hitea} and InternVideo \cite{wang2022internvideo}. For LLM-based models, we use SOTA models such as BLIP-2 \cite{li2023blip}, LLaMA-VQA \cite{ko2023large}, LSTP \cite{wang2024lstp}, SeVILA \cite{yu2023self}, VidF4~\cite{liang2024end}, ViLA~\cite{wang2023vlap} and MotionEpic~\cite{fei2024video}. Among these models, both LLaMA-VQA and MotionEpic use 7B-parameter LLM as part of the model.

\paragraph{Implemententation} For the \textbf{RG} phrase, we use InternVL(v1.5)~\cite{chen2024far, chen2024internvl} as the reasoning generator, 
since it is currently the most powerful open-source MLLM. It has strong reasoning capabilities and matches the performance of commercial closed-source models such as GPT-4V, GPT-4O, and Gemini Pro~\cite{team2023gemini, openai2024gpt4o,openai2024gpt4v} across various benchmarks. We uniformly sample $N$ ($N$ = 4 for faster generation here) frames from the video, then feed these frames to InternVL(26B) to generate complete reasoning processes for each dataset's training set via prompts.
% which is one of SOTA and open-source MLLMs. Its 26B version can also be deploy on a single H800 GPU.

For the \textbf{LR} phrase, we use BLIP-FlanT5~\cite{li2023blip} as our model. Specifically, we employ ViT-G \cite{fang2023eva} as the visual encoder and initialize FlanT5 \cite{chung2022scaling} (3B parameters) as the LLM. We only finetune the modality projection layers and LoRA weights of LLM~\cite{hu2021lora} during training. We use this setting by default for experiments unless otherwise specified.
See Appendix~\ref{app:data} and~\ref{app:train_details} for more details.
% \xjc{LoRA is used but it does not mention?}

\subsection{Model Performance Evaluation}

\paragraph{Overall Reasults}
Tables~\ref{tb:main results} and~\ref{tb:main results-intentqa} provide a comprehensive comparison of our method (\model, 3B) with existing methods across multiple benchmarks.
In Table~\ref{tb:main results}, which presents performance on the NExT-QA and STAR, our model consistently surpasses other methods. \ljx{Our model achieves a total accuracy of 77.0\% on NExT-QA and 74.5\% on STAR, significantly outperforming other models. Notably, compared to methods that also utilize the 3B Flan-T5, our approach demonstrates a clear performance advantage, exceeding them by +2.9 and +7.3, respectively. \model even surpasses models such as LLaMA-VQA, MotionEpic, and Vamos, which rely on larger LLMs as answer generators.} This highlights the robust capability of \model in tackling complex VideoQA tasks.

Table~\ref{tb:main results-intentqa} further illustrates results on the IntentQA. Similar to the previous results in Table~\ref{tb:main results}, \ljx{our model achieves an impressive total accuracy of 77.0\%, surpassing SOTA models by a significant margin, outperforming them by +5.9 (77.0\% vs. 71.1\%), even when they utilize larger LLMs.} \model excels across all question types, with particularly strong performances in the 'How' (93.3\%) and 'Tem.' (69.1\%) categories. This demonstrates \model's effectiveness in handling a diverse range of question types and its superior capability in understanding and generating accurate answers.

The results above demonstrate the effectiveness of our approach, particularly in leveraging reasoning processes to enhance video understanding. Our model consistently achieves significantly better performance across various benchmarks, underscoring its robustness and versatility in VideoQA tasks. The detailed improvements in specific categories further validate the strengths of our methodology.

\paragraph{Adapting \model to Different Model Architectures} \ljx{In addition to integrating \model with encoder-decoder architecture LLMs like FlanT5, we further adapt \model to decoder-only architecture LLMs to demonstrate its generalizability and robustness. Specifically, we apply \model to LLaVA-OV~\cite{li2024llava}, one of the SOTA models in MLLMs that owns strong performance across various tasks. We conduct experiments using two different model scales of LLaVA-OV, namely LLaVA-OV (0.5B) and LLaVA-OV (7B), to validate our method. 
% The results are shown in Table~\ref{tb:llava5}.
}

The results in Table~\ref{tb:llava5} clearly demonstrate the effectiveness of integrating our method across different model architectures and scales. 
\model consistently achieves better performance across models of varying 
\begin{table}[!thb]
\centering
\resizebox{\linewidth}{!}{
\begin{tabular}{l|ccc|c}
\toprule
\multirow{1}{*}{Model}  

&\multirow{1}{*}{NExT-QA}  &\multirow{1}{*}{STAR}  &\multirow{1}{*}{IntentQA} &\multirow{1}{*}{Avg. $\uparrow$}
\\ 
 
 \midrule
% 4b nextqa
% tem: 6044 10774 0.560980137
% cau	10886	17944	0.606665181
% des	3464	5414	0.639822682
% total  0.597503809
% 
% 4b star
% Interaction: 9508 16303 0.5832055449917193
% Sequence: 14086 22262 0.6327374000539036
% Prediction: 2341 4155 0.5634175691937425
% Feasibility: 1634 3011 0.5426768515443374
% total 0.602851457435875
% 
% 4b intentqa
% CW: 3701 5818 0.63612
% TP&N 1476 2664 0.55405
% CH: 1060 1598 0.66332
% total: 0.61875
% 
InternVL(4B) &59.8  &60.3  &61.9 &60.7 \\
InternVL(26B) &67.0  &69.8  &69.0 & 68.6   \\
\bottomrule

\end{tabular}
}
\caption{InternVL's accuracies on three VideoQA  tasks.}
\label{tb:internvl}
\end{table}
sizes, with an average improvement of +2.5 and +2.3 for LLaVA-OV (0.5B) and LLaVA-OV (7B), respectively, across three tasks. \xj{Results indicate that our method is widely and easily adapted to different model architectures, and it scales well with larger models, making it suitable for both smaller and more complex architectures.}

\subsection{Data Impact Analysis}

% \begin{figure*}[!thb]
%     \centering

%     \begin{minipage}[c]{0.6\textwidth}
%     \centering
%     % \includegraphics[scale=0.41]{0_pictures/example-test11.pdf}
%     \includegraphics[width=\textwidth]{0_pic/example.pdf}
%     \end{minipage}
    
%     \hfill
    
%     \begin{minipage}[c]{0.2\textwidth}
%     \centering    \includegraphics[width=1\textwidth]{0_pic/data_eff_nextqa_v2_separate_legends.pdf}
%     \vspace{1em} 
%     \includegraphics[width=1\textwidth]{0_pic/data_eff_nextqa_v2_separate_legends.pdf}
    
%     \end{minipage}
    
%     \caption{An example of a response generated by an MLLM. The sections with an orange background represent the original output, where the generated answer 'a2' is incorrect. Although the answer is wrong, some of the reasoning steps may contain valuable learning elements for the model. For instance, the portions highlighted with a black background provide a partial description of the video and use the process of elimination to rule out two options, offering meaningful insights.}
%     \label{fig:example}
%     % \vspace{-4mm}

% \end{figure*}

\begin{figure*}[htbp]
    \centering
    % 左侧图，占50%宽度，顶部对齐
    \begin{minipage}[c]{0.66\textwidth}
    % \begin{subfigure}[c]{0.6\textwidth}
        \centering
        \includegraphics[width=\textwidth]{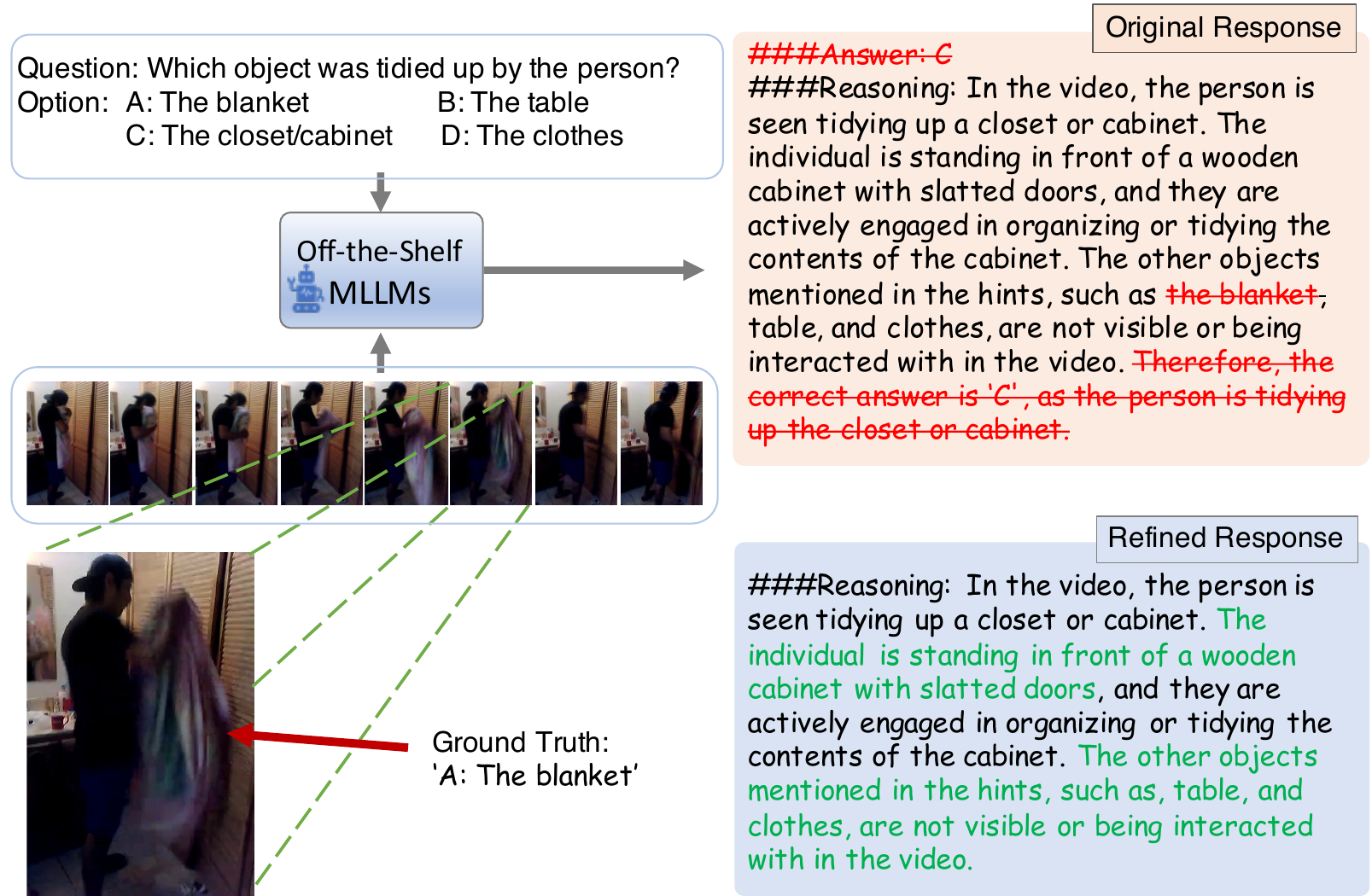}
        \caption{An example of a response generated by an MLLM: the predicted final answer is ``C'', while the true answer is ``A: The blanket''. Although the final answer is incorrect, some of the reasoning steps still offer valuable learning elements for the model. For instance, sentences highlighted in green provide a partial description of the video and eliminate the possibility of two other options, providing meaningful insights.}
        \label{fig:example}
    \end{minipage}
    \hfill
    \begin{minipage}[c]{0.01\textwidth}
        \centering
        \begin{tikzpicture}
            \draw[dashed] (0,0.) -- (0,10.);  % 竖直虚线，长度为5个单位
        \end{tikzpicture}
    \end{minipage}
    % 右侧图，占50%宽度，顶部对齐    
    \begin{minipage}[c]{0.3\textwidth}
        \centering
     % \subcaption{NExT-QA}
     \includegraphics[width=\textwidth]{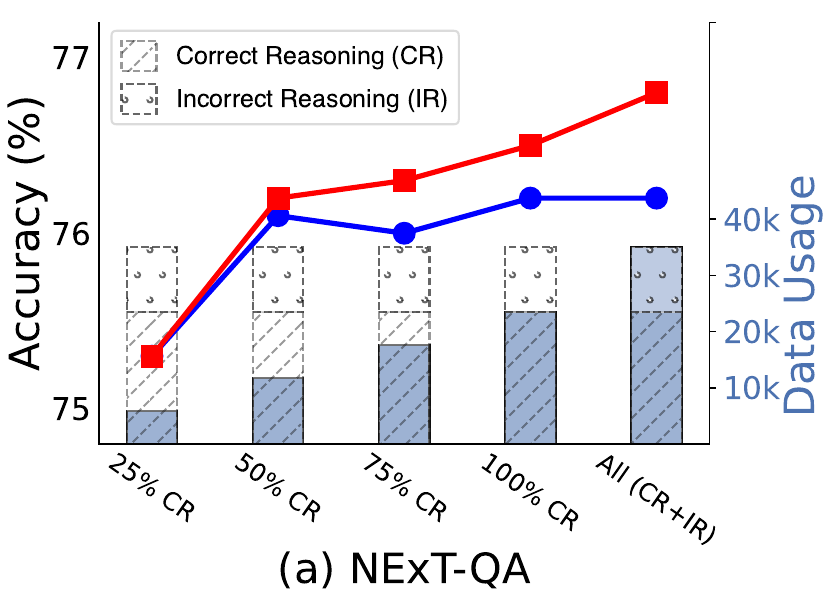}
     \\
     % \subcaption{STAR}
     \includegraphics[width=\textwidth]{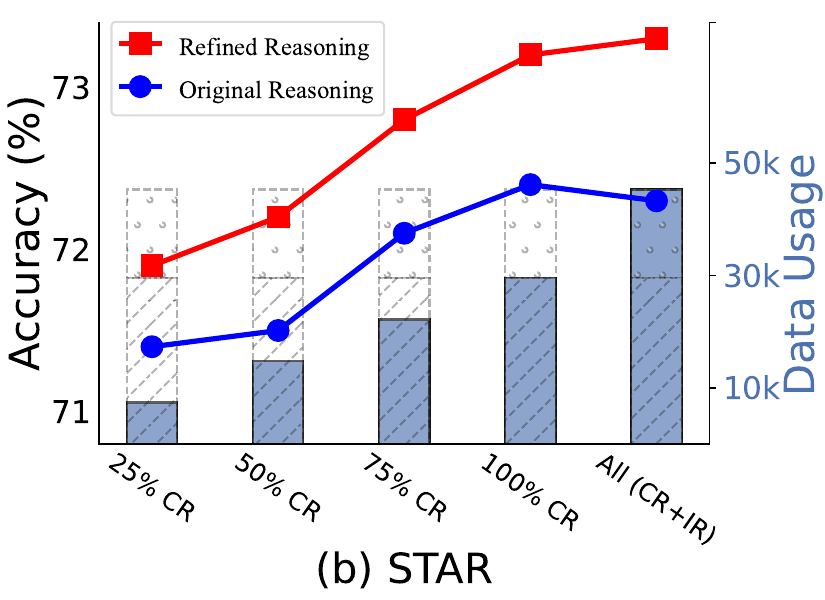}
     % \captionsetup{labelformat=simple}
    \caption{The impact of data quantity on performance and the comparison of Original and Refined Reasoning. Experiments are conducted using the multi-task learning approach, with \(\alpha = \beta = 0.5\).}
    \label{fig:data_effect}
    \end{minipage}
        
    % \end{subfigure}

\end{figure*}

\paragraph{Effectiveness of Reasoning Refinement} \label{sec:data_analysis}
Even though existing MLLMs are strong, when used as zero-shot reasoning generators, they inevitably introduce errors in the reasoning steps due to inherent limitations.
To mitigate the potential impact of such errors in the reasoning, we propose to process and refine these reasoning steps. To further explore it, we conduct experiments validating the effect of data refinement on model performance.

\paragraph{Setup} We categorize the generated reasoning into Correct Reasoning (\textbf{CR}) and Incorrect Reasoning (\textbf{IR}) as described in Section~\ref{sec:rg}. The proportion of CR serves as an indicator of reasoning accuracy, and the results are shown in Table~\ref{tb:internvl}. For example, on NExT-QA, InternVL(26B) achieves an answer accuracy of 67\%, indicating that 33\% of the generated reasoning steps are flawed. In our experimental setup, we test \model trained using both CR and \textbf{All}, where \textbf{All} refers to using all data without distinguishing between CR and IR. For CR, we further experiment with using subsets of 25\%, 50\%, and 75\% of the CR data. 
% The results are presented in Figure~\ref{fig:data_effect}.

% \input{2_tab/ablation_lora}

The results shown in Figure~\ref{fig:data_effect} illustrate the impact of using refined reasoning versus original reasoning on the overall performance across a varying amount of reasoning. For NExT-QA, the performance using original reasoning shows minor fluctuations, with accuracy starting at 75.3\% and slightly increasing to 76.2\% as more correct reasoning data is used. In contrast, the refined reasoning demonstrates a steady upward trend, culminating in an accuracy of 76.8\%, highlighting the effectiveness of reasoning refinement in enhancing performance consistently across different data usage levels. Similarly, the results on STAR exhibit the same trends. 

% Similarly, for STAR, the accuracy with original data starts at 71.4\% and gradually increases to 72.4\%, showing modest improvements. However, when using refined reasoning, the performance starts higher at 71.9\% and consistently improves, reaching 73.3\% at full data usage. This pattern confirms that data refinement significantly boosts performance, especially when larger portions of correct reasoning are utilized.

% \input{2_tab/internvl26}

Based on the results, we can derive the following conclusions:
(1) \textbf{refined reasoning consistently outperforms original reasoning }: In both datasets, the refined reasoning line shows a clear upward trajectory, indicating that refining reasoning substantially improves the model's performance. This effect is more pronounced as the amount of reasoning data increases. (2) \textbf{greater benefits at higher data usage levels}: While improvements are observed at lower levels (e.g., 25\% and 50\%), the differences become more substantial at 75\% and 100\%. This suggests that refinement is especially beneficial when a larger amount of reasoning steps is used, effectively leveraging the available correct data to enhance overall accuracy. (3) \textbf{plateau effect in original reasoning}: For both datasets, using the original reasoning without refinement results in performance that plateaus or slightly fluctuates at higher levels, implying that the presence of incorrect reasoning in the data might counteract the potential benefits of increased data volume. 

\ljx{We can draw consistent conclusions when using InternVL(4B) as the reasoning generator. For more experiments, see Appendix~\ref{app:internvl4}.}

% 

% For additional ablation studies and an exploration of various linguistic materials, please refer to the Appendix~\ref{app:ablation}.

% Overall, these findings highlight the importance of data refinement in improving model performance, especially as the quantity of reasoning data increases. The refined data consistently shows higher accuracy, reinforcing the value of removing incorrect reasoning to better utilize the strengths of the remaining logical flow.

% \textbf{Conclusion 1:} Incorporating reasoning steps improves overall model performance, which aligns with the expectation that "adding more data should lead to performance gains":

\subsection{Methodological Insights}

\begin{figure*}[htbp]
    \centering
    % 左侧图，占50%宽度，顶部对齐
    \begin{minipage}[c]{0.566\linewidth} 
    \centering
    \subfigure[NExT-QA]{
    \includegraphics[width=0.48\textwidth]{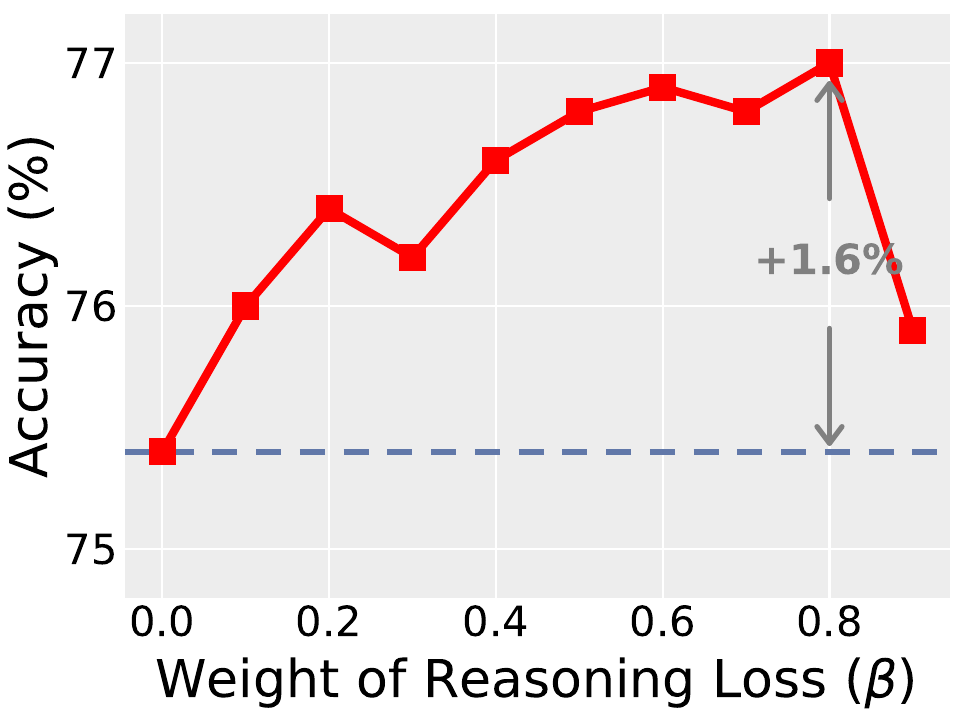}}
    \hfill
    \subfigure[STAR]{
    \includegraphics[width=0.45\textwidth]{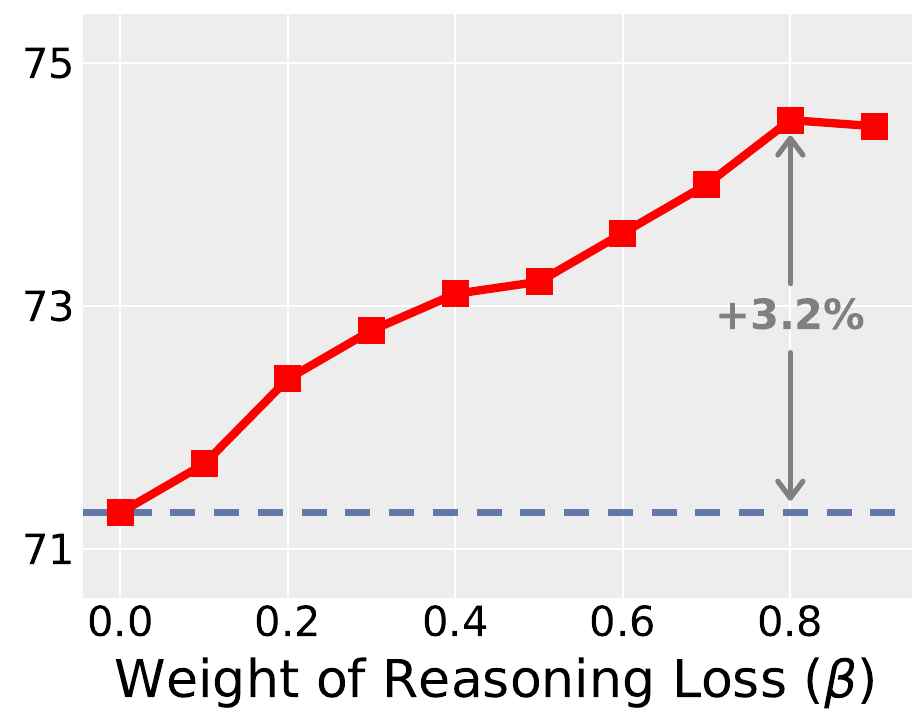}}
    \caption{Hyper-parameter tuning for the weight \(\beta\) of the Reasoning Generation Loss $\mathcal{C}_{rea}(\cdot)$, with the corresponding \(\alpha = 1 - \beta\). }
    \label{fig:hyper_beta}
    \end{minipage}
    % \hfill
    % 右侧图，占50%宽度，顶部对齐    
    \hfill
    \begin{minipage}[c]{0.01\textwidth}
        \centering
        \begin{tikzpicture}
            \draw[dashed] (0,0.) -- (0,5.);  % 竖直虚线，长度为5个单位
        \end{tikzpicture}
    \end{minipage}
    \begin{minipage}[c]{0.4\textwidth}
    \centering
     % \subcaption{NExT-QA}
     \resizebox{\linewidth}{!}{
     \begin{tabular}{l|ccc|c}
     \toprule
     \multirow{2}{*}{Setting}  

     &\multirow{2}{*}{\textbf{NExT-QA}}  &\multirow{2}{*}{\textbf{STAR}}  &\multirow{2}{*}{\textbf{IntentQA}}
     &\multirow{2}{*}{\textbf{Avg.\ $\uparrow$}}\\
     & & & \\ 
     \midrule
     STL$_{QA}$ &75.4  &71.0  &73.3  &73.2  \\
     \midrule
     STL$_{CR}$&74.2  &71.9  &74.4 &73.5  \\
     STL$_{All}$  &73.3  &70.6  &70.8  &71.6 \\
     MTL$_{CR}$ &76.2  &72.9  &76.1  & 75.1 \\
     MTL$_{All}$ &76.8  &73.2  &77.0  & 75.7 \\
     \bottomrule
     \end{tabular}}
     % \captionsetup{labelformat=simple}
     \captionof{table}{The impact of different training approaches on the model performance. All experiments are conducted with \(\alpha = \beta =0.5\) for MTL setups.}
     \label{tb:mt_learning}
     \end{minipage}
        
    % \end{subfigure}

\end{figure*}

\paragraph{Hyper-parameter Tuning}
To increase flexibility in model training, we introduce weighting parameters $\alpha$ and $\beta$ into our framework, as shown in Equation~\ref{eq:loss_mt}, to control the influence of reasoning processes in the model's training objective. We perform a hyperparameter search to evaluate the model’s performance as the weight \(\beta\) for the reasoning generation cross-entropy loss is varied.

Figure~\ref{fig:hyper_beta} presents the results on NExT-QA and STAR. When \(\beta = 0\), only the QA loss contributes to training, serving as a baseline for comparison. The results clearly show that integrating reasoning generation loss improves model performance. Specifically, on NExT-QA, accuracy steadily increases as \(\beta\) rises from 0 to 0.8, with accuracy peaking at 77.0\% compared to the baseline of 75.4\%, which suggests that incorporating the reasoning generation loss helps the model better understand video-based questions by enhancing contextual comprehension and reasoning capabilities. Performance stabilizes between \(\beta = 0.5\) and \(\beta = 0.8\), indicating an optimal balance between reasoning generation and QA objectives. However, when \(\beta\) exceeds 0.8, performance slightly declines to 75.9\% at \(\beta = 0.9\), indicating that overemphasizing reasoning generation may negatively impact VideoQA performance. 
Similarly, results on STAR exhibit the same trends and conclusions.
% Similarly, on the STAR dataset, accuracy increases from 71.3\% at \(\beta = 0\) to 74.5\% at \(\beta = 0.8\), with the most significant improvements occurring between \(\beta = 0.7\) and \(\beta = 0.8\). However, beyond this point, performance plateaus and slightly drops to 74.3\% at \(\beta = 0.9\), further reinforcing the idea that excessive weighting of reasoning can lead to diminishing returns.

% \input{2_tab/ablation_info}

Overall, these results demonstrate the importance of balancing reasoning generation and QA training objectives. Notably, this flexibility offers substantial benefits when carefully managed, allowing the model to better leverage the strengths of both reasoning and question answering. Additionally, it enables customization based on different datasets or task complexities, further improving the model's adaptability and overall performance.

\paragraph{Single Task or Multi-Task?} 
% We perform an ablation study to assess the impact of multi-task learning with reasoning on three benchmarks. As shown in Table~\ref{tb:mt_learning},
% To further understand the benefits of multi-task learning and the impact of using reasoning data in training, we compare different training settings involving the inclusion of both correct and incorrect reasoning processes.

To further understand the impact of different training approaches, as mentioned in Section~\ref{sec:training}, we compare single-task learning (STL) and multi-task learning (MTL). In the STL setting, we evaluate three types of supervision signals: (1) using only the original true answer, denoted as $_{QA}$, (2) concatenating the true answer with the Correct Reasoning, meaning only samples with Correct Reasoning are used, denoted as $_{CR}$, and (3) concatenating all samples with their corresponding reasoning, denoted as $_{All}$.

The results are shown in Table~\ref{tb:mt_learning}. In the STL setting, we observe a minor performance improvement of +0.3 (73.2\% vs. 73.5\%) when correct reasoning is concatenated with the true answer. However, performance drops significantly by +1.6 (73.2\% vs. 71.6\%) when all reasoning samples, correct or not, are used. Conversely, in the MTL setting, using correct reasoning achieves a higher performance of 75.1\%. After refining the reasoning, incorporating all samples leads to further improvements, reaching a performance of 75.7\%.

% These findings suggest that MTL is more robust when integrating reasoning processes, particularly after refining the data, whereas STL can suffer from performance degradation when incorporating noisy reasoning data.

These results highlight several key insights regarding the effectiveness of incorporating reasoning data:
(1) Incorrect reasoning, when used in single-task training, can significantly degrade performance, indicating the importance of filtering incorrect data.
(2) Multi-task learning consistently outperforms single-task setups, as it enables the model to better leverage reasoning processes, even when both correct and incorrect reasoning are present.
(3) Joint learning of QA and reasoning generation provides substantial gains, highlighting the importance of multi-task learning for VideoQA.

See Appendix~\ref{app:ablation} for the abltion of each component and the impact of various linguistic materials.

\subsection{Complex Reasoning Tasks}
To further evaluate the complex reasoning ability of our method, we conduct experiments using the ATP-hard subset~\cite{buch2022revisiting} of the NExT-QA validation set, which is for more challenging causal and temporal questions.  This subset filters out those
'easy' questions that can be answered with a single frame. The results are in Table~\ref{tb:main results-atp}.

These results show that our method improves over VideoAgent~\cite{wang2025videoagent} by 9.3\% on this challenging subset, demonstrating its ability to enhance complex reasoning tasks and mitigate potential issues related to single-frame biases.
\begin{table}[!thb]
\centering
\resizebox{\linewidth}{!}{
\begin{tabular}{c|ccl}
\toprule
\multirow{1}{*}{Model}  

&\multirow{1}{*}{Cau.}  &\multirow{1}{*}{Tem.}  &\multirow{1}{*}{Tot. $\uparrow$} 
\\ 
 
 \midrule
% 4b nextqa
% tem: 6044 10774 0.560980137
% cau	10886	17944	0.606665181
% des	3464	5414	0.639822682
% total  0.597503809
% 
% 4b star
% Interaction: 9508 16303 0.5832055449917193
% Sequence: 14086 22262 0.6327374000539036
% Prediction: 2341 4155 0.5634175691937425
% Feasibility: 1634 3011 0.5426768515443374
% total 0.602851457435875
% 
% 4b intentqa
% CW: 3701 5818 0.63612
% TP&N 1476 2664 0.55405
% CH: 1060 1598 0.66332
% total: 0.61875
% 

Temporal[ATP]~\cite{buch2022revisiting} &38.4 &36.5 &38.8 \\ 
GF~\cite{bai2024glance} &48.7 & 50.3& 49.3 \\
VideoAgent~\cite{wang2025videoagent} &57.8  &58.8  &58.4  \\
\midrule
BLIP-FlanT5 (w. ReasVQA) &\textbf{69.2 } &\textbf{65.7}  &\textbf{67.7}(+9.3)    \\
\bottomrule

\end{tabular}
}
\caption{{Model comparison on ATP-hard subset.}
% \huzhangc{It is important to record the model size in the table.}
}
\label{tb:main results-atp}
\end{table}

\section{Conclusion}

In this paper, we introduce \textbf{\model} (Reasoning-enhanced Video Question Answering), a novel approach that utilizes reasoning processes generated by MLLMs to improve the performance of smaller VideoQA models. 
% Our method comprises two phases: Reasoning Generation (RG) and Learning From Reasoning (LFR). 
Through extensive experiments on three benchmarks, \model achieves new SOTA results, demonstrating the effectiveness of leveraging reasoning processes in enhancing video understanding. Our in-depth analysis validates each step of \model, highlighting the value of incorporating refined reasoning data and multi-task learning to enhance the model’s capabilities. Results indicate that by guiding models with intermediate reasoning steps, we can significantly boost performance, particularly on complex reasoning tasks. Future work will explore further refinement strategies and integration processes, aiming to extend the approach to other multimodal tasks.

\section*{Limitations}
Our method is straightforward but relies on the quality of reasoning generated by MLLMs. Although the performance of the VideoQA model improves significantly after refinement, the incorrect or biased reasoning produced by these MLLMs can still negatively affect the overall performance. In future work, we will continue to investigate how to generate higher-quality reasoning or develop better refinement strategies to enable the model to extract more consistent and meaningful information from the reasoning data.

\section*{Acknowledgements}

The authors thank the kind suggestions and support from AI Data Technology Laboratory of Huawei Noah's Ark Lab.

\bibliography{naacl_latex}

\appendix

\section{Experiments Setup}
\label{app:setup}
\subsection{Dataset Details}
\label{app:data}

\textbf{NExT-QA}~\cite{xiao2021next} is a VideoQA benchmark targeting the explanation of video content. The video in NExT-QA primarily encompasses aspects of daily life, social interactions, and outdoor activities, featuring three types of questions: \textit{Temporal} (Tem), \textit{Causal} (Cau), and \textit{Descriptive} (Des). It contains 5.4k videos and about 52K manually annotated question-answer pairs, each QA pair comprises one question and five candidate answers.

\textbf{STAR}~\cite{wu2021star} is oriented towards real-world reasoning scenarios, encompassing four question types, namely \textit{Interaction} (Int), \textit{Sequence} (Seq), \textit{Prediction} (Pre), and \textit{Feasibility} (Fea). STAR contains 22K Situation Video Clips and 60K Situated Questions.

\textbf{IntentQA}~\cite{li2023intentqa} is a special kind of inference VideoQA dataset that focuses on intent reasoning which studies inference VideoQA beyond factoid VideoQA. 

% \textbf{TVQA} is a multi-choice QA benchmark containing 152k QA pairs and 21k video clips from 460 hours of video. It features videos from genres such as sitcoms, medical dramas, and crime series. 

\begin{table}[h]
\centering
% \renewcommand{\thetable}{5}

% \resizebox{\linewidth}{!}{
\begin{tabular}{c|ccc}
\toprule
\multirow{1}{*}{} & 
\multicolumn{1}{c}{\textbf{NExT-QA}}  &\multicolumn{1}{c}{\textbf{STAR}}  &\multicolumn{1}{c}{\textbf{IntentQA}} \\

 \midrule
 
\rowcolor{white!10}\#videos &5.4k  &22k  &4.3k      \\
\rowcolor{white!20}\#questions &52k  &60k  &16k    \\

\bottomrule
\end{tabular}
% }
\caption{Statistics of the datasets we used.}
\label{tab:statistics}
\end{table}

% \section{Training hyper-parameters}

% We conducted experiments using 4 $\times$ A100 (80G) GPUs. For each dataset, including NExT-QA~\cite{xiao2021next}, STAR~\cite{wu2021star}, and TVQA~\cite{lei2018tvqa}, the data batch size per GPU was set to 8, with a learning rate of 3e-5. We implemented a warm-up of 1000 steps and trained the model for 10 epochs.

\subsection{Experimental Details}

\paragraph{Training details}\label{app:train_details}
We finetune the modality projection layers and LLM(LoRA)~\cite{hu2021lora} during training in NVIDIA H800(80GB) GPU $\times$ 1. We use AdamW with a cosine learning rate scheduler, whose max learning rate is 3e-5, and a batch size of 8, We train our model within 10 epochs. Our training code is implemented based on LAVIS~\footnote{\url{https://github.com/salesforce/LAVIS}} and transformers~\footnote{\url{https://github.com/huggingface/transformers}} libraries, and will be later open sourced.

% \textbf{BLIP-2} processes all frames by voting or concatenating them and then uses LLM to generate the final answer.

\paragraph{Baselines}\label{app:baseline} We compare our method with two types of baselines: non-LLM and LLM-based models. For non-LLM methods, we use recent SOTA models, including Just Ask \cite{yang2021just}, All-in-One \cite{wang2023all} and MIST \cite{gao2023mist}, HiTeA \cite{ye2023hitea} and InternVideo \cite{wang2022internvideo}. For LLM-based models, we use SOTA models such as BLIP-2 \cite{li2023blip}, LLaMA-VQA \cite{ko2023large}, LSTP \cite{wang2024lstp}, SeVILA \cite{yu2023self}, VidF4~\cite{liang2024end}, ViLA~\cite{wang2023vlap}, Vamos~\cite{wang2023vamos} and MotionEpic~\cite{fei2024video}. Among these models, LLaMA-VQA, Vamos, and MotionEpic use 7B-parameter LLM as part of the model.

\textbf{LLaMA-VQA} is built based on LLaMA-7B \cite{touvron2023llama}, enabling the model to understand the complex relationships between videos, questions, and answers by constructing multiple auxiliary tasks.

\textbf{LSTP} adopts the BLIP-2 architecture and uses optical flow for frame selection, followed by using LLM to generate answers. 

\textbf{SeVILA} relies on a multi-stage training process and is trained on an additional dataset with temporal localization supervision. During the inference, it first utilizes BLIP-2 and LLMs for frame selection and then uses BLIP-2 and LLM again for answer generation. 

\textbf{MotionEpic} breaks down the raw intricate video reasoning problem into a chain of simpler sub-problems and solves them one by one sequentially.

\textbf{Vamos}~\cite{wang2023vamos} generalizes the concept bottleneck model to work with tokens and nonlinear models, which uses hard attention to select a small subset of tokens from the free-form text as inputs to the LLM reasoner.

% \section{Implementation Details}
% \label{app: implementation}
% We use ViT-G as the visual encoder and Flan-T5 XL as the large language model. For the frame selector, we employ a Transformer encoder structure as the question-answer encoder. The text encoder of QFS and the multimodal fusion module (QFM) share parameters, both initialized with the parameters of Qformer pretrained in the first stage.
% We use BLIP-2 as our answer generator. Specifically, we employ ViT-G \cite{fang2023eva} as the visual encoder and initialize FlanT5-XL \cite{chung2022scaling} (3B parameters) as the LLM. We use the Qformer trained in the \textbf{second-stage} of BLIP-2 to connect ViT and FlanT5-XL, which is trained on a large dataset of image-text pairs to align visual representations to the representation space of the LLM.
% For the feature extractor \texttt{E$_e$($\cdot$)} and text encoder \texttt{E$_t$($\cdot$)} in QFS, as well as the multimodal fusion encoder \texttt{E$_f$($\cdot$)} in QFM, we employ the Transformer encoder architecture. \texttt{E$_e$($\cdot$)}, \texttt{E$_t$($\cdot$)} and \texttt{E$_f$($\cdot$)} share parameters (188M parameters), and they are all initialized with the parameters of Qformer pretrained in the \textbf{first-stage} of BLIP-2. 

\subsection{Prompts used by MLLMs.}\label{app:prompts}

We use prompt engineering to employ MLLMs to generate reasoning processes. We present the prompt template as follows:

~

\noindent"""

\noindent\texttt{These frames are uniformly sampled from a video. Given a question about the video, you should choose the correct answer option from a list of possible answers based on the video content and respond with the option in the format `\#\#\#Answer: A'. You should also provide a detailed reasoning process explaining why the chosen answer is correct. Cite specific details from the video frames to support your answer. Explain each step of the reasoning to ensure that the answer is logical and reliable. \#\#\#Question: {question}, \#\#\#Hints: {options}.}"""

\subsection{Reasoning Refinement Details}\label{app:rr}
To address the conflict between process supervision and final result supervision during training, specifically targeting inconsistencies between generated reasoning and true answers, we apply reasoning refinement to the original reasoning generated by MLLMs. Our reasoning refinement phase removes any sentences in the OR that contain the conclusion or predicted answer. Specifically,  we identify and remove the sentences with the following fixed patterns:

\noindent"""

\noindent\texttt{'\#\#\#Answer: ... '}

\noindent\texttt{'**Answer**: ... '}

\noindent\texttt{'\#\#\#Conclusion: ... '}

\noindent\texttt{'**Conclusion**: ... '}

\noindent\texttt{'\#\#\#Detailed Explanation ... '}

\noindent\texttt{'The correct answer ... '}

\noindent\texttt{'Thus, the correct answer is ... '}

\noindent\texttt{'Therefore, the correct answer is ... '}

\noindent\texttt{'Based on these observations ... '}

\noindent\texttt{'Given these observations and the context ... '}"""

% \noindent\texttt{' ... '"""}

This ensures the reasoning focuses on the process rather than the final conclusion, reducing potential bias.

% \section{Experiments}
\begin{table}[!thb]
\centering
\resizebox{\linewidth}{!}{
\begin{tabular}{c|ccccc}
\toprule
\multirow{2}{*}{Setting}  &
\multicolumn{5}{c}{\textbf{STAR}} \\
&\textbf{Int.}  &\textbf{Seq.}  &\textbf{Pre.}  &\textbf{Fea.}  &\textbf{Tot.}  \\ 
\midrule
 \multicolumn{1}{l|}{\model (3B)}  &75.9  &76.6  &67.3  &62.0  &74.5  \\
\rowcolor{gray!10}  \multicolumn{1}{l|}{w/o. Reasoning}  &71.4  &73.5  &63.0  &62.7  &71.0    \\
\rowcolor{gray!10} \multicolumn{1}{l|}{w/o. LoRA}  
&68.5  &71.1  &59.3  &58.0  &68.3   
\\
\rowcolor{gray!10} \multicolumn{1}{l|}{w/o. both} &65.1  &69.3  &58.7  &58.0  &66.0   \\
\bottomrule
\end{tabular}
}
\caption{Ablation study for each component of \model.} 
\label{tb:ablation_lora}
\end{table}

\section{More Anylysis}\label{app:ablation}
% We conduct the
% ablation experiments on two additional datasets in Table~\ref{app:ablation}. 
\subsection{Ablations}\label{app:component}
\paragraph{Ablation Study for Each Component of ReasVQA}
To investigate the effect of each component in our framework, we conduct an extensive ablation study. 
Table~\ref{tb:ablation_lora} presents an ablation study evaluating the impact of various components on model performance, specifically focusing on the reasoning process (Rea) and Low-Rank Adaptation (LoRA). Our model, \model (3B), which integrates both reasoning processes and LoRA, achieves a Total accuracy of 74.5. This demonstrates the effectiveness of incorporating these elements, significantly improving performance over the reasoning processes alone.

When the reasoning process is omitted from \model, the Total accuracy decreases to 71.0. This drop highlights the significant contribution of the reasoning process to the model's overall performance. In the absence of LoRA, the model's Total accuracy is 68.3, showing that while LoRA improves performance, its effect is less pronounced compared to the reasoning process.
The removal of both the reasoning process and LoRA results in the lowest Total accuracy of 66.0. This underscores the combined importance of both components, as their exclusion notably impairs the model's effectiveness.

Overall, the ablation study indicates that both the reasoning process and LoRA are crucial for optimal performance. The reasoning process has a more substantial impact, while LoRA also contributes positively but to a lesser degree. The highest performance is achieved when both components are utilized, demonstrating their complementary roles in enhancing the model's capabilities.

\begin{table}[!thb]
\centering
\resizebox{\linewidth}{!}{
\begin{tabular}{l|ccccc}
\toprule
\multirow{2}{*}{Setting}  &
\multicolumn{5}{c}{\textbf{STAR}} \\
&\textbf{Int.}  &\textbf{Seq.}  &\textbf{Pre.}  &\textbf{Fea.}  &\textbf{Tot.}  \\ 
 
\midrule

\multicolumn{1}{c|}{InternVL (26B)} &66.0  &68.9  &63.8  &62.2  &67.0 \\

\midrule
QA only &71.4  &73.5  &63.0  &62.7  &71.0  \\

\midrule
% generate brief caption
\ \ \ w. Cap$_{brief}$  &70.7  &73.1  &64.4  &61.4  &70.7  \\
% generate detailed caption
\ \ \ w. Cap$_{detailed}$ &72.1  &73.7  &65.2  &63.1  &71.7    \\

% generate reasoning process 
\ \ \ w. Reasoning  &74.1  &76.0  &62.3  &61.8  &73.2   \\
\bottomrule

\end{tabular}
}
\caption{Influence of different linguistic materials on \model. All experiments are conducted using a multi-task learning approach, with both \(\alpha\) and \(\beta\) set to 0.5.}
\label{tb:ablation_info}
\end{table}

% \subsection{}\label{app:materials}
\subsection{Effects of Different Linguistic Materials on ReasVQA}\label{app:materials}
% \paragraph{The impact of using different linguistic information.} 
We also evaluate the model's performance when using different types of linguistic information as training data.
Table~\ref{tb:ablation_info} illustrates the results of different types of additional information on the multi-task learning of the model, including settings with QA only, brief captions, detailed captions, and reasoning processes.

The performance of InternVL (26B) is reported with a Total accuracy of 67.0, reflecting the quality of the reasoning processes it produces. InternVL is included not as a baseline for our model, but to provide a reference for the accuracy of the reasoning processes it generates.

The baseline model using only the QA loss function achieves a Total accuracy of 71.0. When brief captions are added, the accuracy across various categories decreases, resulting in a reduced Total accuracy of 70.7. This suggests that brief captions may provide insufficient contextual information, failing to significantly enhance model performance.
In contrast, when detailed captions are used, accuracy improves across all categories, raising the Total accuracy to 71.7. This indicates that detailed captions offer richer contextual information, which contributes to better overall model performance.

When our approach is applied, utilizing reasoning processes generated by a multi-modal model to supervise the small model's reasoning generation, significant improvements are observed in Interaction (74.1) and Sequence (76.0) metrics. The Total accuracy further increases to 73.2, which is a 3.2\% improvement compared to the QA-only setting (73.2 vs. 71.0). These results highlight that reasoning processes significantly enhance the model's understanding and answering capabilities, providing the most substantial performance gains.

\begin{table}[!thb]
\centering
\resizebox{\linewidth}{!}{
\begin{tabular}{c|ccc|ccc}
\toprule

\multirow{2}{*}{Setting}  &
\multicolumn{3}{c|}{\textbf{NExT-QA$\uparrow$}} &\multicolumn{3}{c}{\textbf{IntentQA$\uparrow$}} \\
&\textbf{75\% CR} &\textbf{CR}  &\textbf{All}  &\textbf{75\% CR} &\textbf{CR}  &\textbf{All}  \\ 
\midrule
 
Original &75.0  &75.9	&75.5 &75.0 &75.7	&75.7	 \\
\midrule
Refined &75.9 &76.2	&76.5 &76.3 &76.3  &76.4   \\
% \midrule
% \multicolumn{6}{c}{\textbf{IntentQA}}\\
% \midrule
% Original    &75	&75	&75	&75.7	&75.7  \\

% Refined  &75.1  &75.4	&76.3	&76.4	&76.5    \\
\bottomrule
\end{tabular}
}
\caption{The impact of Original Reasoning and Refined Reasoning generated by InternVL(4B) on \model.} 
\label{tb:internvl4b}
\end{table}

% \begin{table}[!thb]
% \centering
% \resizebox{\linewidth}{!}{
% \begin{tabular}{c|ccccc}
% \toprule
% \multirow{2}{*}{Setting}  &
% \multicolumn{5}{c}{\textbf{IntentQA}} \\
% &\textbf{25\%}  &\textbf{50\%}  &\textbf{75\%}  &\textbf{100\%}  &\textbf{All}  \\ 
% \midrule
 
% Original   &75.0 &75.0 &75.0 &75.7 &75.7  \\

% Refined    &75.1 &75.4 &76.3 &76.4 &76.5    \\
% \midrule
% \end{tabular}
% }
% \caption{The impact of Original Reasoning and Refined Reasoning generated by InternVL(4B) on \model.} 
% \label{tb:internvl4b}
% \end{table}

\subsection{Impact of Reasoning Generated by InternVL(4B)}\label{app:internvl4}
% \paragraph{Impact of Reasoning Generated by InternVL(4B)}
To further evaluate the impact of the reasoning generated by the MLLMs (reasoning generator) on our model, we employ InternVL-4B to generate reasoning processes. Following the methods outlined in Sections~\ref{sec:rg} and~\ref{sec:data_analysis}, we categorize the generated reasoning into "Correct Reasoning" and "Incorrect Reasoning." 
% From the correct reasoning data, we randomly select 25%, 50%, 75%, and 100%, while "All" represents the indiscriminate use of both "Correct" and "Incorrect Reasoning." The results are presented in Table~\ref{tb:ablation-reasoning}.
It is evident that when using all reasoning data, performance does not surpass that of using only correct reasoning; in fact, it is lower, which highlights the negative impact of incorrect reasoning on model training. After applying refinement, performance improves across all categories, especially with 75\% correct reasoning, where the model shows a 1.3\% improvement compared to the original reasoning. This indicates that refinement is beneficial, even for correct reasoning.

Importantly, after refinement, using all data (both correct and incorrect) achieves higher performance than using only correct reasoning. This suggests that, while incorrect reasoning may contain erroneous conclusions, it still offers some meaningful information. 

Additionally, when combining the results from Figure~\ref{fig:data_effect}, Table~\ref{tb:internvl}, and Table~\ref{tb:internvl4b}, we find that the stronger the MLLM, the higher the accuracy of the reasoning it generates. Moreover, incorporating more correct reasoning consistently enhances model performance.

% \subsection{Test on the NExT-QA ATP Hard Set}
% To evaluate the complex reasoning ability of our method, we conduct experiments by using NExT-QA ATP Hard Set~\cite{buch2022revisiting}, which is for more challenging causal and temporal questions.

\end{document}